\documentclass[runningheads]{llncs}
\usepackage{graphicx}
\usepackage{booktabs}
\usepackage{amsfonts}
\usepackage{subfigure}
\usepackage[linesnumbered,ruled,vlined]{algorithm2e}

\newcommand\orcidicon[1]{\href{https://orcid.org/#1}{\mbox{\scalerel*{
\begin{tikzpicture}[yscale=-1,transform shape]
\pic{orcidlogo};
\end{tikzpicture}
}{|}}}}

\begin{document}
\title{UTNet: A Hybrid Transformer Architecture for Medical Image Segmentation}
%
%
\author{Yunhe Gao  \inst{1}  \and
Mu Zhou\inst{1,2} \and
Dimitris Metaxas\inst{1}
}
\authorrunning{Y. Gao et al.}
%
\institute{Department of Computer Science, Rutgers University \and
SenseBrain and Shanghai AI Laboratory and Centre for Perceptual and Interactive Intelligence
\\
}
\maketitle              
\begin{abstract}
Transformer architecture has emerged to be successful in a number of natural language processing tasks. However, its applications to medical vision remain largely unexplored. In this study, we present UTNet, a simple yet powerful hybrid Transformer architecture that integrates self-attention into a convolutional neural network for enhancing medical image segmentation. UTNet applies self-attention modules in both encoder and decoder for capturing long-range dependency at different scales with minimal overhead. To this end, we propose an efficient self-attention mechanism along with relative position encoding that reduces the complexity of self-attention operation significantly from $O(n^2)$ to approximate $O(n)$. A new self-attention decoder is also proposed to recover fine-grained details from the skipped connections in the encoder. Our approach addresses the dilemma that Transformer requires huge amounts of data to learn vision inductive bias. Our hybrid layer design allows the initialization of Transformer into convolutional networks without a need of pre-training. We have evaluated UTNet on the multi-label, multi-vendor cardiac magnetic resonance imaging cohort. UTNet demonstrates superior segmentation performance and robustness against the state-of-the-art approaches, holding the promise to generalize well on other medical image segmentations. Code is available\footnote{https://github.com/yhygao/UTNet}.
\end{abstract}
%
%
\section{Introduction}




Convolutional networks have revolutionized the computer vision field with outstanding feature representation capability. Currently, the convolutional encoder-decoder architectures have made substantial progress in position-sensitive tasks, like semantic segmentation \cite{ronneberger2015u,isensee2021nnu,wang2017central,tajbakhsh2020embracing,gao2019multi}. The used convolutional operation captures texture features by gathering local information from neighborhood pixels. To aggregate the local filter responses globally, these models stack multiple convolutional layers and expand the receptive field through down-samplings. Despite the advances, there are two inherent limitations of this paradigm. First, the convolution only gathers information from neighborhood pixels and lacks the ability to capture long-range (global) dependency explicitly \cite{zhao2017pyramid,yu2015multi,gao2021focusnetv2}. Second, the size and shape of convolution kernels are typically fixed thus they can not adapt to the input content \cite{schlemper2019attention}.

Transformer architecture using the self-attention mechanism has emerged to be successful in natural language processing (NLP) \cite{vaswani2017attention} with its capability of capturing long-range dependency. Self-attention is a computational primitive that implements pairwise entity interactions with a context aggregation mechanism, which has the ability to capture long-range associative features. It allows the network to aggregate relevant features dynamically based on the input content. Preliminary studies with simple forms of self-attention have shown its usefulness in segmentation \cite{fu2019dual,sinha2020multi}, detection \cite{YI2019228} and reconstruction \cite{huang2019mri}.


\begin{figure}[t]
\setlength{\abovecaptionskip}{5pt} 
\setlength{\belowcaptionskip}{-10pt}

\includegraphics[width=\linewidth]{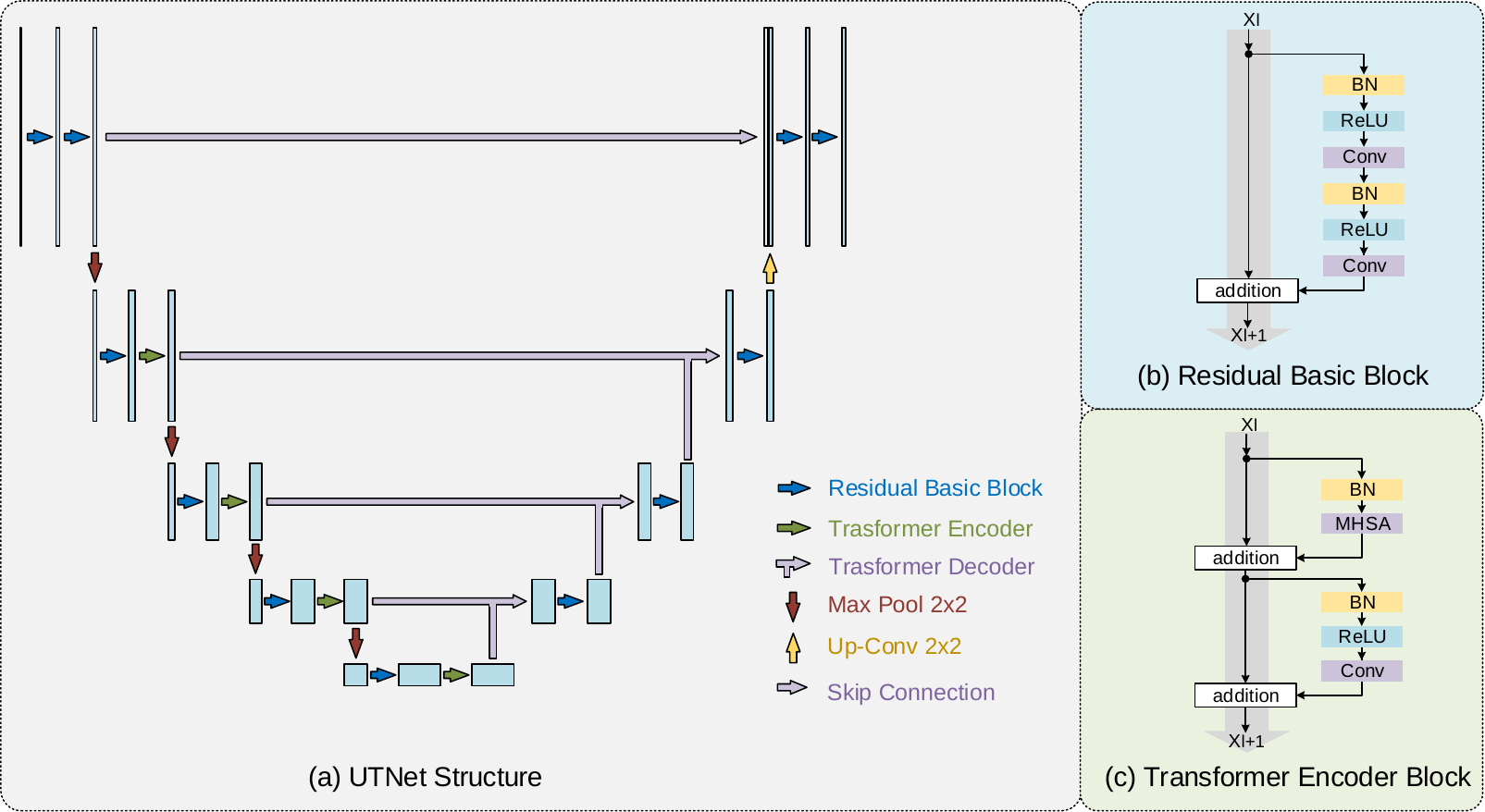}
\caption{\textbf{(a)} The hybrid architecture of the proposed UTNet. The proposed efficient self-attention mechanism and relative positional encoding allow us to apply Transformer to aggregate global context information from multiple scales in both encoder and decoder. \textbf{(b)} Pre-activation residual basic block. \textbf{(c)} The structure of Transformer encoder block.  }\label{framework}
\setlength\tabcolsep{2pt}
\end{figure}

Although the application of image-based Transformer is promising, training and deploying of Transformer architecture has several daunting challenges. First, the self-attention mechanism has $O(n^2)$ time and space complexity with respect to sequence length, resulting in substantial overheads of training and inference. Previous works attempt to reduce the complexity of self-attention \cite{huang2019ccnet,zhu2019asymmetric}, but are still far from perfection. Due to the time complexity, the standard self-attention can be only applied  patch-wise, e.g. \cite{dosovitskiy2020image,zheng2020rethinking} encode images using $16\times16$ flattened image patches as input sequences, or on top of feature maps from CNN backbone, which are already down-sampled into low-resolution \cite{fu2019dual,wang2018non}. However, for position-sensitive tasks like medical image segmentation, high-resolution feature plays a vital role since most mis-segmented areas are located around the boundary of the region-of-interest. Second, Transformers do not have inductive bias for images and can not perform well on a small-scale dataset \cite{dosovitskiy2020image}. For example, Transformer can be beneficial from pre-training through a large-scale dataset like full JFT-300M \cite{dosovitskiy2020image}. But even with pre-training on ImageNet, Transformer is still worse than the ResNet \cite{kolesnikov2019big,he2016deep}, not to mention medical image datasets with much less available amounts of medical data.

In this paper, we propose a \textbf{U}-shape hybrid \textbf{T}ransformer \textbf{Net}work: \textbf{UTNet}, integrating the strength of convolution and self-attention strategies for medical image segmentation. The major goal is to apply convolution layers to extract local intensity features to avoid large-scale pretraining of Transformer, while using self-attention to capture long-range associative information. We follow the standard design of UNet, but replace the last convolution of the building block in each resolution (except for the highest one) to the proposed Transformer module. Towards enhanced quality of segmentation, we seek to apply self-attention to extract detailed long-range relationships on high-resolution feature maps. To this end, we propose an efficient self-attention mechanism, which reduces the overall complexity significantly from $O(n^2)$ to approximate $O(n)$ in both time and space. Furthermore, we use a relative position encoding in the self-attention module to learn content-position relationships in medical images. Our UTNet demonstrates superior segmentation performance and robustness in the multi-label, multi-vendor cardiac magnetic resonance imaging cohort. Given the design of UTNet, our framework holds the promise to generalize well on other medical image segmentations.

\section{Method}

\subsection{Revisiting Self-attention Mechanism}

The Transformer is built upon the multi-head self-attention (MHSA) module \cite{vaswani2017attention}, which allows the model to jointly infer attention from different representation subspaces. The results from multiple heads are concatenated and then transformed with a feed-forward network. In this study, we use 4 heads and the dimension of multi-head is not presented for simplicity in the following formulation and in the figure. Consider an input feature map $X\in \mathcal{R}^{C\times H\times W}$, where $H$,$W$ are the spatial height, width and $C$ is the number of channels. Three $1\times 1$ convolutions are used to project $X$ to query, key, value embeddings: $\textbf{Q, K, V}\in \mathcal{R}^{d\times H\times W}$, where $d$ is the dimension of embedding in each head.  The $\textbf{Q, K, V}$ is then flatten and transposed into sequences with size $n\times d$, where $n=HW$. The output of the self-attention is a scaled dot-product:

\begin{equation}
    {\rm Attention}(\textbf{Q, K, V})=\underbrace{{\rm softmax}(\frac{\textbf{QK}^{\mathsf{T}}}{\sqrt{d}})}_{P}\textbf{V}
\end{equation}

Note that $P\in \mathcal{R}^{n\times n}$ is named context aggregating matrix, or similarity matrix. To be specific, the $i$-th query's context aggregating matrix is $P_i={\rm softmax}(\frac{\mathbf{q}_i\mathbf{K}^\mathsf{T}}{\sqrt{d}})$, $P_i\in \mathcal{R}^{1\times n}$, which computes the normalized pair-wise dot production between $q_i$ and each element in the keys. The context aggregating matrix is then used as the weights to gather context information from the values. In this way, self-attention intrinsically has the global receptive field and is good at capturing long-range dependence. Also, the context aggregating matrix is adaptive to input content for better feature aggregation. However, the dot-product of $n\times d$ matrices leads to $O(n^2d)$ complexity. Typically, $n$ is much larger than $d$ when the resolution of feature map is large, thus the sequence length dominates the self-attention computation and makes it infeasible to apply self-attention in high-resolution feature maps, e.g. $n=256$ for $16\times 16$ feature maps, and $n=16384$ for $128\times 128$ feature maps.

\begin{figure}[t]
\setlength{\abovecaptionskip}{5pt} 
\setlength{\belowcaptionskip}{-13pt}

\includegraphics[width=\linewidth]{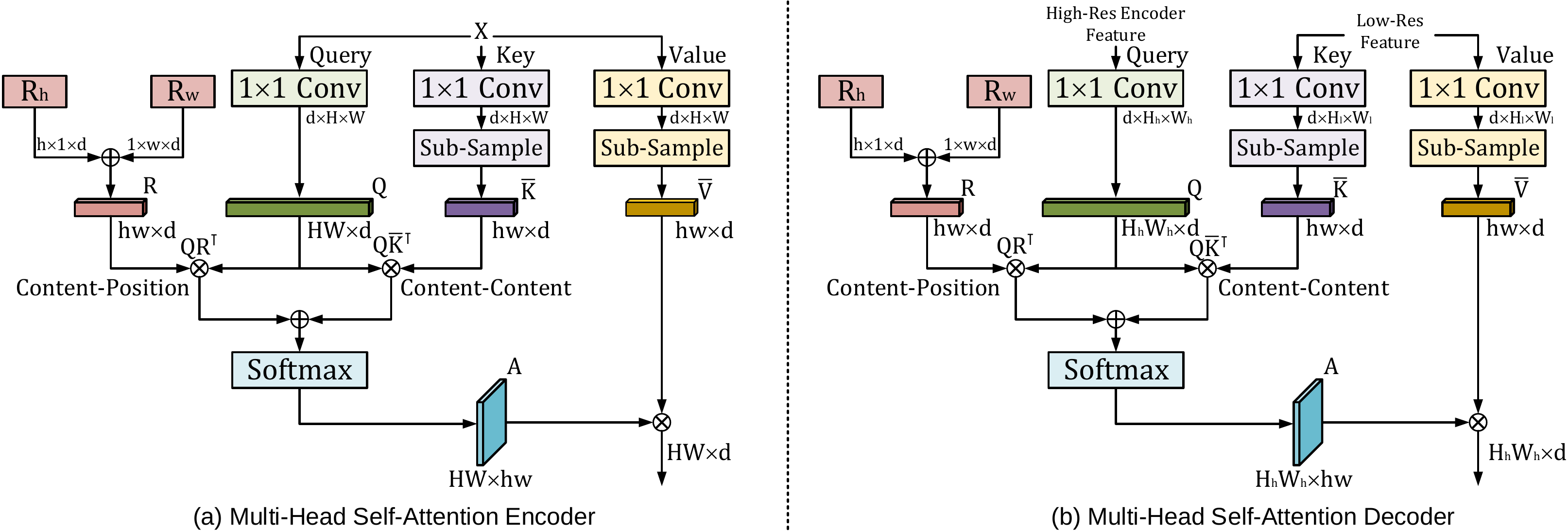}
\caption{The proposed efficient multi-head self-attention (MHSA). \textbf{(a)} The MHSA used in the Transformer encoder. \textbf{(b)} The MHSA used in the Transformer decoder. They share similar concepts, but (b) takes two inputs, including the high-resolution features from skip connections of the encoder, and the low-resolution features from the decoder.}\label{sa}
\setlength\tabcolsep{2pt}
\end{figure}

\subsection{Efficient Self-attention Mechanism}

As images are highly structured data, most pixels in high-resolution feature maps within local footprint share similar features except for the boundary regions. Therefore, the pair-wise attention computation among all pixels is highly inefficient and redundant. From a theoretical perspective, self-attention is essentially low rank for long sequences \cite{wang2020linformer}, which indicates that most information is concentrated in the largest singular values. Inspired by this finding, we propose an efficient self-attention mechanism for our task as seen in Fig. \ref{sa}.

The main idea is to use two projections to project key and value: $\mathbf{K, V}\in \mathcal{R}^{n\times d}$ into low-dimensional embedding: $\overline{\mathbf{K}}, \overline{\mathbf{V}}\in \mathcal{R}^{k\times d}$, where $k=hw\ll n$, $h$ and $w$ are the reduced size of feature map after sub-sampling. The proposed efficient self-attention is now:

\begin{equation}
    {\rm Attention}(\textbf{Q}, \overline{\textbf{K}}, \overline{\textbf{V}})=\underbrace{{\rm softmax}(\frac{\textbf{Q}\overline{\textbf{K}}^{\mathsf{T}}}{\sqrt{d}})}_{\overline{P}:n\times k}\underbrace{\overline{\textbf{V}}}_{k\times d}
\end{equation}

By doing so, the computational complexity is reduced to $O(nkd)$. Notably, the projection to low-dimensional embedding can be any down-sampling operations, such as average/max pooling, or strided convolutions. In our implementation, we use $1\times 1$ convolution followed by a bilinear interpolation to down-sample the feature map, and the reduced size is 8.

\subsection{Relative Positional Encoding}

Standard self-attention module totally discards the position information and is perturbation equivariant \cite{bello2019attention}, making it ineffective for modeling image contents that are highly structured. The sinusoidal embedding in previous works \cite{parmar2018image} does not have the property of translation equivariance in convolutional layers. Therefore, we use the 2-dimensional relative position encoding by adding relative height and width information \cite{bello2019attention}. The pair-wise attention logit before softmax using relative position encoding  between pixel $i=(i_x,i_y)$ and pixel $j=(j_x,j_y)$ :
\begin{equation}
    l_{i,j}=\frac{q_i^\mathsf{T}}{\sqrt{d}}(k_j+r_{j_x-i_x}^{W}+r_{j_y-i_y}^{H})
\end{equation}

where $q_i$ is the query vector of pixel $i$, $k_i$ is the key vector for pixel $j$, $r_{j_x-i_x}^{W}$ and $r_{j_y-i_y}^{H}$ are learnable embeddings for relative width $j_x-i_x$ and relative height $j_y-i_y$ respectively. Similar to the efficient self-attention, the relative width and height are computed after low-dimensional projection. The efficient self-attention including relative position embedding is:

\begin{equation}
    {\rm Attention}(\textbf{Q}, \overline{\textbf{K}}, \overline{\textbf{V}})=\underbrace{{\rm softmax}(\frac{\textbf{Q}\overline{\textbf{K}}^{\mathsf{T}}+\textbf{S}_H^{rel}+\textbf{S}_W^{rel}}{\sqrt{d}})}_{\overline{P}:n\times k}\underbrace{\overline{\textbf{V}}}_{k\times d}
\end{equation}

where $\textbf{S}_H^{rel}, \textbf{S}_W^{rel}\in\mathcal{R}^{HW\times hw}$ are matrics of relative position logits along height and width dimensions that satisfy $\textbf{S}_H^{rel}[i,j]=q_i^{\mathsf{T}}r^H_{j_y-i_y}, \textbf{S}_W^{rel}[i,j]=q_i^{\mathsf{T}}r^W_{j_x-i_x}$.

\subsection{Network Architecture}
Fig. 1 highlights the architecture of UTNet. We seek to combine the strength from both convolution and self-attention mechanism. Therefore, the hybrid architecture can leverage the inductive bias of image from convolution to avoid large-scale pretraining, as well as the capability of Transformer to capture long-range relationships. Because the mis-segmented region usually locates at the boundary of region-of-interest, the high-resolution context information could play a vital role in segmentation. As a result, our focus is placed on the proposed self-attention module, making it feasible to handle large-size feature maps efficiently. Instead of naively integrating the self-attention module on top of the feature maps from the CNN backbone, we apply the Transformer module to each level of the encoder and decoder to collect long-range dependency from multiple scales. Note that we do not apply Transformer on the original resolution, as adding Transformer module in the very shallow layers of the network does not help in experiments but introduces additional computation. A possible reason is that the shallow layers of the network focus more on detailed textures, where gathering global context may not be informative.
The building block of UTNet is shown in Fig. \ref{framework} (b) and (c), including residual basic block and Transformer block. For both blocks, we use the pre-activation setting for identity mapping in the short cut. This identity mapping has been proven to be effective in vision \cite{he2016identity} and NLP tasks \cite{wang2019learning}.

\section{Experiment}

\subsection{Experiment Setup}
We systematically evaluate the UTNet on the multi-label, multi-vendor cardiac magnetic resonance imaging (MRI) challenge cohort \cite{mmchallenge}, including the segmentation of left ventricle (LV), right ventricle (RV), and left ventricular myocardium (MYO). In the training set, we have 150 annotated images from two different MRI vendors (75 images of each vendor), including A: Siemens; B: Philips. In the testing set, we have 200 images from 4 different MRI vendors (50 images of each vendor), including A: Siemens; B: Philips; C: GE; D: Canon, where vendor C and D are completely absent in the training set (we discard the unlabeled data). The MRI scans from different vendors have marked differences in appearance, allowing us to measure model robustness and compare with other models under different settings. Specifically, we have performed two experiments to highlight the performance and robustness of UTNet. First, we report primary results with training and testing data are both from the same vendor A. Second, we further measure the cross-vendor robustness of models. Such setting is more challenging since the training and testing data are from independent vendors. We report Dice score and Hausdorff distance of each model to compare the performance.

\subsection{Implementation Detail}

For data preprocessing, we resample the in-plane spacing to $1.2\times 1.2$ mm, while keeping the spacing along the z-axis unchanged. We train all models from scratch for 150 epochs. We use the exponentially learning rate scheduler with a base learning rate of 0.05. We use the SGD optimizer with a batch size of 16 on one GPU, momentum and weight decay are set to 0.9 and $1e-4$ respectively. Data augmentation is applied on the fly during model training, including random rotation, scaling, translation, additive noise and gamma transformation. All images are randomly cropped to $256\times 256$ before entering the models. We use the combine of Dice loss and cross-entropy loss to train all networks.

\begin{table}[tb]
\centering
 \scriptsize
\setlength\tabcolsep{10pt}

\caption{Segmentation performance in term of Dice score and efficiency comparison. All models are trained and tested using data from vendor A. The Hausdorff distant result is reported in the supplementary.}\label{indomain}
\begin{tabular}{l|c|c|c|c|c}
\toprule
      & UNet   & ResUNet &CBAM &Dual-Attn &UTNet\\
\hline
LV    &91.8     &92.2   &92.2   &92.4   &\textbf{93.1}  \\
MYO   &81.7     &82.5   &82.1   &82.3   &\textbf{83.5}  \\
RV    &85.6     &86.2   &87.7   &86.4   &\textbf{88.2}  \\
\hline
Average   &86.4     &86.9   &87.3   &87.0   & \textbf{88.3}       \\\hline
Params/M & 7.07   &9.35 &9.41 & 9.69 &9.53\\ \hline
Inference Time/s & 0.085   & 0.115 &0.149 & 0.118 &0.145\\
\bottomrule
\end{tabular}
\end{table}

\subsection{Segmentation Results}

We compare the performance of UTNet with multiple state-of-the-art segmentation models. UNet \cite{ronneberger2015u} builds on top of the fully convolutional networks with a U-shaped architecture to capture context information. The ResUNet is similar to UNet in architecture, but it uses residual blocks as the building block. CBAM \cite{woo2018cbam} uses two sequential convolutional modules to infer channel and spatial attention to refine intermediate feature maps adaptively. Dual attention network \cite{fu2019dual} uses two kinds of self-attention to model the semantic inter-dependencies in spatial and channel dimensions, respectively. We have implemented CBAM and dual attention in ResUNet backbone for better comparison. The dual attention is only applied in the feature maps after 4 down-samplings due to its quadratic complexity.

As seen in Table \ref{indomain}, UTNet demonstrates leading performance in all segmentation outcomes (LV, MYO and RV). By introducing residual connections, ResUNet is slightly improved than the original UNet. The spatial and channel attention from CBAM are inferred from convolutional layers, it still suffers from limited receptive field. Thus CBAM only has limited improvement compared with ResUNet. We also recognize that dual-attention approach was almost the same as ResUNet, as it suffers from quadratic complexity that can not process higher resolution feature maps to fix errors in the segmentation boundary. Meanwhile, our UTNet presents less parameters than dual-attention approach and it can capture global context information from high-resolution feature maps.


%




\begin{figure}[tb]

\includegraphics[width=\linewidth]{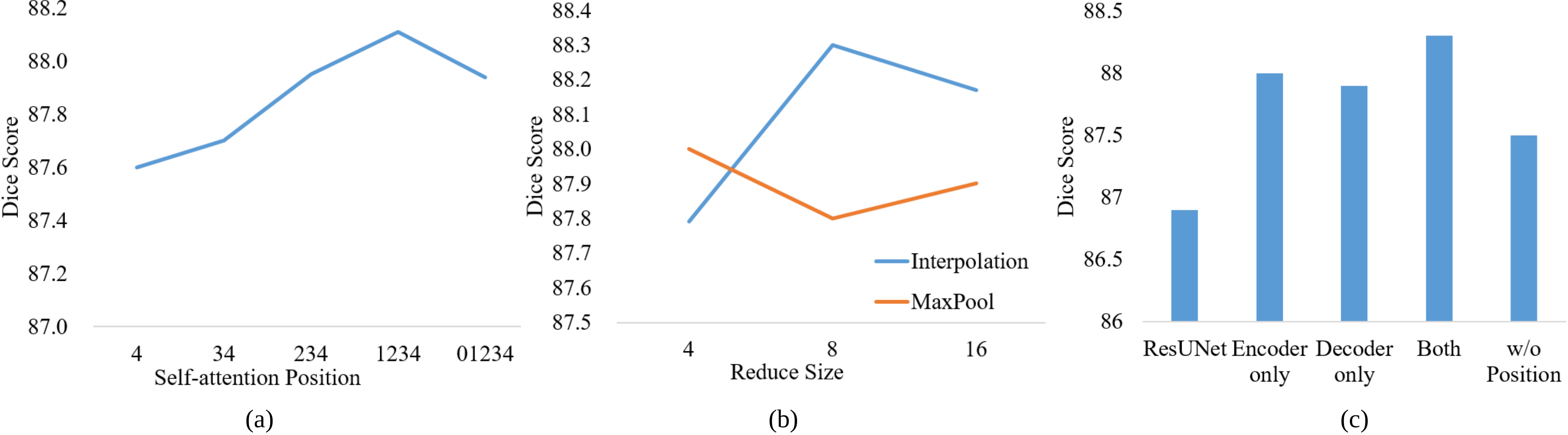}
\caption{Ablation study. \textbf{(a)} Effect of different self-attention position. \textbf{(b)} Effect of reduce size and projection of efficient self-attention. \textbf{(c)} Effect of Transformer encoder, Transformer decoder, and the relative positional encoding. }\label{ablation}
\setlength\tabcolsep{2pt}
\end{figure}

\begin{table}[tb]
\centering

 \tiny
\setlength\tabcolsep{2pt}

\caption{Robustness comparison, measured with Dice score. All models are trained on data from vendor A,B, and are tested on data from vendor A,B,C,D. The number in brackets of C and D indicates the performance drop compared with the average of A and B. }\label{robustness}
\begin{tabular}{l|c|c|c|c|c|c|c|c|c|c|c|c}
\toprule
      & \multicolumn{4}{|c|}{ResUNet} &  \multicolumn{4}{|c|}{CBAM} &  \multicolumn{4}{|c}{UTNet}   \\
      \hline
Vendor &A &B &C &D &A &B &C &D &A &B &C &D\\
\hline
LV      & 92.5 & 90.1 & 88.7 ($\downarrow$2.6) & 87.2 ($\downarrow$4.1) & \textbf{93.3} & 91.0 & 89.4 ($\downarrow$2.8) & 88.8 ($\downarrow$3.4) & 93.1 &\textbf{91.4} & \textbf{89.8 ($\downarrow$2.5)} & \textbf{90.5 ($\downarrow$1.8)}    \\
\hline
MYO   & 83.6 & 85.3 & 82.8 ($\downarrow$1.7) & 80.2 ($\downarrow$4.3) & \textbf{83.9} & 85.8 & 82.6 ($\downarrow$2.3) & 80.8 ($\downarrow$4.1) & 83.7 & \textbf{85.9} & \textbf{83.7 ($\downarrow$1.1)}& \textbf{82.6 ($\downarrow$2.2)}\\
\hline
RV  & 87.4 & 87.5 & 85.9 ($\downarrow$1.6) & 85.3 ($\downarrow$2.2) & 88.4 & 88.4 & 85.3 ($\downarrow$3.1)& 86.4 ($\downarrow$2.0) & \textbf{89.4} & \textbf{88.8} & \textbf{86.3 ($\downarrow$2.8)} & \textbf{87.3 ($\downarrow$1.8)} \\
\hline
AVG    &87.9 & 87.6 & 85.7 ($\downarrow$2.0) & 84.2 ($\downarrow$3.5) & 88.5 & 88.4 & 85.5 ($\downarrow$2.7)& 85.3 ($\downarrow$3.2)& \textbf{88.7} & \textbf{88.7} &\textbf{86.6 ($\downarrow$2.1)}& \textbf{86.2  ($\downarrow$2.5)}\\
\bottomrule
\end{tabular}
\end{table}

\begin{figure}[tbh]
\centering

\includegraphics[width=0.8\linewidth]{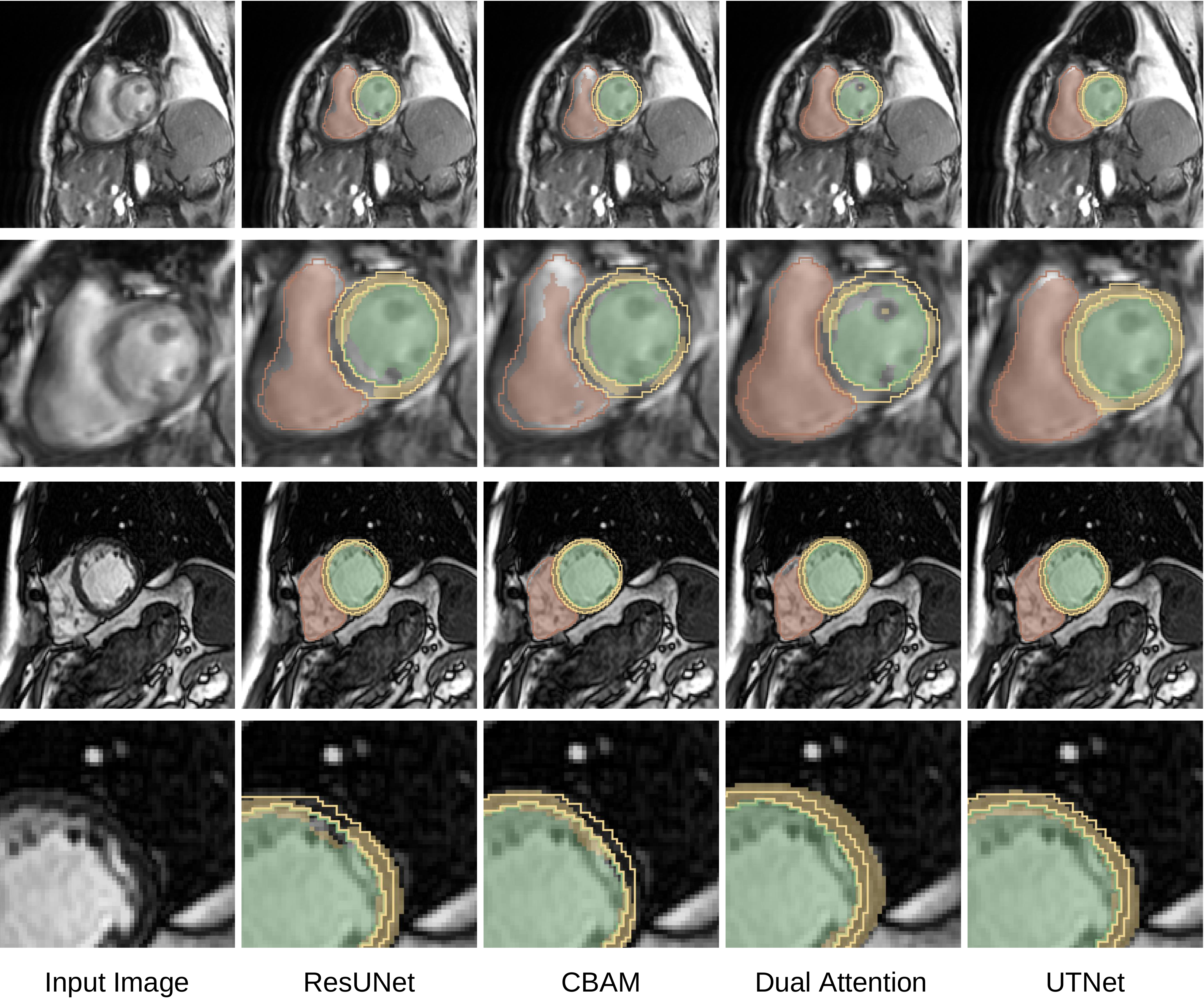}
\caption{Hard cases visualization on unseen testing data from vendor C and D. First two rows and the bottom two rows present the results and a zoom-in view of vendor C and D, respectively. The outline indicates the ground-truth annotation. Best viewed in color with LV(green), MYO(yellow), and RV(red). The test case from vendor C is blur due to motion artifacts, while the test case from vendor D is noisy and has low contrast in the boundary. Only UTNet provides consistent segmentation, which demonstrates its robustness. More visualization of segmentation outcomes are presented in the supplementary. }\label{seg_vis}
\setlength\tabcolsep{2pt}
\end{figure}

\noindent
\textbf{Ablation study}. Fig. \ref{ablation} (a) shows the performance of different self-attention positions. The number in the x-axis indicates the level where self-attention is places, e.g., '34' means the level where 3 and 4 times down-samplings are performed. As the level goes up, the self-attention can gather more fine-grained detail information with increased performance. However, the curve saturates when adding to the original resolution. We reason this as the very shallow layer tends to be more focused on local texture, where global context information is not informative anymore. Fig. \ref{ablation} (b) shows the result of efficient self-attention's reduced size of 4, 8, 16. The reduced size 8 results in the best performance. The interpolation down-sampling is slightly better than using max-pooling. Fig. \ref{ablation} (c) shows the effect of the Transformer encoder, decoder, and the relative positional encoding using the optimal hyper-parameter from (a) and (b). The combination of the Transformer encoder and decoder gives the optimal performance. The relative positional encoding also plays a vital role, as removing it causes a large performance drop.

For a head-to-head comparison with standard self-attention on space and time complexity, we further apply dual attention in four resolutions (1, 2, 3, 4, same as UTNet), and use the same input image size and batch size ($256\times256\times16$) to test the inference time and memory consumption. UTNet gains superior advantage over dual attention with quadratic complexity, where GPU memory: 3.8 GB vs 36.9 GB and time: 0.146 s vs 0.243 s.

\noindent
\textbf{Robustness analysis}. Table \ref{robustness} shows results on training models with data from vendor A and B, and then test the models on vendor A, B, C, and D, respectively. When viewing results on C and D vendors, competing approaches suffer from vendor differences while UTNet retains competitive performance. This observation can probably be attributed to the design of self-attention on multiple levels of feature maps and the content-position attention, allowing UTNet to be better focused on global context information instead of only local textures. Fig. 4 further shows that UTNet displays the most consistent results of boundaries, while the other three methods are unable to capture subtle characteristics of boundaries, especially for RV and MYO regions in cardiac MRI.

\section{Conclusion}

We have proposed a U-shape hybrid Transformer network (UTNet) to merge advances of convolutional layers and self-attention mechanism for medical image segmentation. Our hybrid layer design allows the initialization of Transformer into convolutional networks without a need of pre-training. The novel self-attention allows us to extend operations at different levels of the network in both encoder and decoder for better capturing long-range dependencies. We believe that this design will help richly-parameterized Transformer models become more accessible in medical vision applications. Also, the ability to handle long sequences efficiently opens up new possibilities for the use of the UTNet on more downstream medical image tasks.


\subsubsection{Acknowledgement}

This research was supported in part by NSF: IIS 1703883, NSF IUCRC CNS-1747778 and funding from SenseBrain, CCF-1733843, IIS-1763523, IIS-1849238, MURI- Z8424104 -440149 and NIH: 1R01HL127661-01 and R01HL127661-05. and in part by Centre for Perceptual and Interactive Intellgience (CPII) Limited, Hong Kong SAR.

%
%
%
%
\bibliographystyle{splncs04}
\bibliography{egbib}

\end{document}